\documentclass[conference, 9pt]{IEEEtran}
\IEEEoverridecommandlockouts
\usepackage{cite}
\usepackage{amsmath,amssymb,amsfonts}
\usepackage{algorithmic}
\usepackage{graphicx}
\usepackage{textcomp}
\usepackage{xcolor}
\usepackage{caption}
\usepackage{subcaption}
\usepackage{threeparttable}
\usepackage{url}

\usepackage{breakurl}
\usepackage[breaklinks]{hyperref}
\usepackage[numbers,sort&compress]{natbib}

\usepackage{enumitem}

\def\BibTeX{{\rm B\kern-.05em{\sc i\kern-.025em b}\kern-.08em
    T\kern-.1667em\lower.7ex\hbox{E}\kern-.125emX}}
    
\newcommand{\mysubsec}[1]{
\noindent\textbf{#1}}

\begin{document}

\title{
\vspace{-12pt}
Software/Hardware Co-design for Multi-modal Multi-task Learning in Autonomous Systems
}

\author{\IEEEauthorblockN{Cong Hao}
\IEEEauthorblockA{
Georgia Institute of Technology\\
callie.hao@ece.gatech.edu}
\and
\IEEEauthorblockN{Deming Chen}
\IEEEauthorblockA{University of Illinois at Urbana-Champaign\\
dchen@illinois.edu}
}
\maketitle

\begin{abstract}
Optimizing the quality of result (QoR) and the quality of service (QoS) of AI-empowered autonomous systems simultaneously is very challenging.
First, there are multiple input sources, e.g., multi-modal data from different sensors, requiring diverse data preprocessing, sensor fusion, and feature aggregation. Second, there are multiple tasks that require various AI models to run simultaneously, e.g., perception, localization, and control. Third, the computing and control system is heterogeneous, composed of hardware components with varied features, such as embedded CPUs, GPUs, FPGAs, and dedicated accelerators. 
Therefore, autonomous systems essentially require multi-modal multi-task (MMMT) learning which must be aware of hardware performance and implementation strategies.
While MMMT learning has been attracting intensive research interests, its applications in autonomous systems are still underexplored. In this paper, we first discuss the opportunities of applying MMMT techniques in autonomous systems, and then discuss the unique challenges that must be solved.
In addition, we discuss the necessity and opportunities of MMMT model and hardware co-design, which is critical for autonomous systems especially with power/resource-limited or heterogeneous platforms. 
We formulate the MMMT model and heterogeneous hardware implementation co-design as a differentiable optimization problem, with the objective of improving the solution quality and reducing the overall power consumption and critical path latency. 
We advocate for further explorations of MMMT in autonomous systems and software/hardware co-design solutions.
\end{abstract}

\section{Introduction}

Many machine learning assisted autonomous systems, such as autonomous driving cars, drones, robotics, warehouse and factory systems, are essentially \textit{multi-modal multi-task (MMMT) learning with dedicated hardware requirements and implementation}~\cite{Baltrusaitis2019Multimodal, zhang2017survey, chowdhuri2019multinet, aranjuelo2018multimodal, merrill2019end}. First, the system inputs usually involve data from multiple sensors, such as images from multiple cameras, point cloud from Radar and Lidar, ultrasonic signals, data from position, velocity, and acceleration sensors, and sound and motion signals.
This phenomena is called sensor fusion~\cite{luo2011multisensor}, which involves multi-modal learning~\cite{aranjuelo2018multimodal, huang2020multi} when the sensors provide different data modalities.
Second, the system tasks may include sensing, perception, localization, planning, controlling, actuation, etc.; each task may include multiple subtasks, such as object detection, recognition, and segmentation inside perception.
This phenomena is recognized as multi-task learning (MTL)~\cite{zhang2017survey, merrill2019end, akhtar2020deep, lee2020context}.
Third, autonomous systems, especially with embedded devices such as drones and small robots, are often with limited computation capability, memory capacity, and battery budget, but require low latency, low power consumption, and less dependence on the communication links; thus, deliberately designed hardware platforms and accelerators are in pressing demand~\cite{zhang2017current}. 
Given the complexity of inputs, tasks, and hardware platforms, autonomous system design and optimization is very challenging.

Recognizing the opportunities of MMMT learning in autonomous systems, 
several recent works try to optimize the autonomous system designs through MMMT learning technologies~\cite{Yang2018EndtoendMM, Hou2019LearningTS, chowdhuri2019multinet, solomon2019hierarchical}.
% such as MultiNet~\cite{kim2020multi} in autonomous driving and VLocNet++~\cite{radwan2018vlocnet++} in robots.
For instance, MultiNet~\cite{kim2020multi} is an end-to-end autonomous driving model with a main task of predicting motor and steering control and auxiliary tasks consisting of additional inferred speed and steering values, while the inputs include image data, motor, steering, and three distinct behavioral modes.
Another recent work~\cite{Yang2018EndtoendMM} uses a single-task network for steering prediction with a multi-modal network for speed prediction, while the inputs include images and past inferred driving speeds.
Targeting other autonomous systems such as robots, VLocNet++~\cite{radwan2018vlocnet++} is a multitask learning approach that exploits the inter-task relationship between estimating global pose, odometry, and semantic segmentation.

While promising, the exploitation of MMMT in autonomous system is still in the early stage, 
%where the considered inputs and tasks are largely simplified from realistic, 
and there are unique challenges that remain unsolved.
Specifically, to exploit MMMT learning in autonomous systems, the following challenges must be addressed:
\begin{itemize}
    \item {
    \textit{Collaborative modalities.} Sensor fusion in autonomous systems is more challenging than multi-modality, not only because it involves more and varied modalities, but also because it requires harmonizing with different modality features, reconciling with conflicting decisions, and integrating and synthesizing of the global understanding, with an ultimate goal of better understanding the surrounding environment \textit{as a whole}. 
    }
    \item {
    \textit{Collaborative tasks.} One essential difference of the MTL in autonomous systems is that, in many cases, the tasks aim to collaboratively achieve one ultimate goal, rather than achieving high quality in each independent task. For example in autonomous driving, all different tasks serve for the final decision making for steering, accelerating, and breaking. Therefore, the unique question in autonomous systems is how to explore inter-relationships across multi-tasks towards collaborative objectives.
    }
    \item {
    \textit{Hardware deployment matters.} Autonomous systems usually require real-time response (especially for life critical decisions) and/or low power consumption (e.g., drones, robots). 
    Since the MMMT models are powerful yet complex, the question that how the models help or harm hardware deployment must be answered, which becomes more challenging when dealing with heterogeneous computing platforms.
    }
\end{itemize}

To address the challenges, we propose \textit{MMMT model and heterogeneous implementation co-design}, aiming to search for both high QoR software solutions and high QoS deployments. Further, we propose a 
\textit{unified multi-objective formulation} for the co-design problem considering both QoR and QoS and their tradeoffs, which allows the problem to be systematically studied and solved via various optimization techniques such as discrete or continuous optimizations.
We also discuss potential solutions of the co-design problem.

In the remaining sections, we first discuss the related works in Section~\ref{sec:related}, and then discuss the unique challenges and opportunities of applying MMMT in autonomous systems in Section~\ref{sec:collab_mmmt_learning}. In Section~\ref{sec:sw-hw-codesign}, we discuss the opportunities of MMMT and hardware co-design and propose a unified formulation which can be solved as an optimization problem. Section~\ref{sec:conclusion} concludes the paper.

\section{Related Works}
\label{sec:related}

\mysubsec{Multi-modal learning.} 
Data modality refers to the way in which the data are represented, such as image, text, and sound.
Multi-modal machine learning aims to process and relate information from multiple sources to capture the correspondences between modalities and gain an in-depth understanding of natural phenomena, and usually focuses on modality representation, translation, alignment, fusion, and co-learning~\cite{Baltrusaitis2019Multimodal}.
In autonomous systems, sensor fusion is one type of multi-modal learning~\cite{gao2020survey}.
For instance, PointPainting~\cite{Vora2020PointPaintingSF} and FusionLane~\cite{Yin2020FusionLaneMF} both fuse image-only segmentation with lidar points,
while CRF-Net~\cite{Nobis2019ADL} and CenterFusion~\cite{Nabati2020CenterFusionCR} fuses camera data and sparse radar data for object detection.
% Brena et al.~\cite{brena2020choosing} propose to use machine learning to choose the best fusion method from eight fusion configurations.
Huang et al.~\cite{huang2020multi} propose an end-to-end model for multi-modal sensor fusion with visual and depth information of images.

\mysubsec{Multi-task learning (MTL).} 
In MTL~\cite{zhang2017survey}, machine learning models are trained with data from multiple tasks simultaneously, usually using shared weights, aiming to unveil the relations between tasks and to increase training efficiency.
Many studies have demonstrated the success of MTL, such as language processing~\cite{radford2019language}, question answering~\cite{mccann2018natural}, and sentence simplification~\cite{guo2018dynamic}
%smile detection and emotion recognition~\cite{sang2017multi}, and comic book image analysis~\cite{nguyen2019comic}.

\mysubsec{Multi-modal multi-task learning.}
Multi-task learning sometimes also deal with multiple data modalities, yielding MMMT learning,
where the input representations are shared across both modalities and tasks.
A representative work, one model does all, proposed by Kaiser et al.~\cite{kaiser2017one}, is a MMMT model that can handle multiple tasks with varying input domains.
Similarly, Omninet~\cite{pramanik2019omninet} is another unified architecture for MMMT.
Most recently, Lu et al.~\cite{lu202012} introduce a multi-task model that handles twelve different datasets simultaneously.

\mysubsec{Multi-modal and/or multi-task in autonomous systems.}
Given the success of MMMT learning in the joint area of computer vision and natural language processing, there is a great number of multi-modal learning or multi-task learning approaches proposed for autonomous driving~\cite{sistu2019neurall, aranjuelo2018multimodal, unlu2019sliding, kim2020multi, warakagoda2019fusion, Hou2019LearningTS, zheng2020improving}.
For example, NeurAll~\cite{sistu2019neurall} is a three-task model that performs segmentation, depth estimation, and moving object detection at the same time,
while Warakagoda et al.~\cite{warakagoda2019fusion} propose a neural network-based fusion architecture whose inputs are camera images and lidar images.
However, the joint MMMT approaches in autonomous systems are still limited.
MultiNet~\cite{chowdhuri2019multinet} is a recent MMMT learning approach for autonomous driving where the primary task is steering and motor value prediction and the auxiliary task is trajectory prediction.
%the input data include left and right images from the stereo camera and additional information called behavioral modes, including direct, follow, and furtive modes.
Yang et al.~\cite{Yang2018EndtoendMM} propose a MMMT network to predict the steering angles and speeds simultaneously by taking previous feedback speeds as an extra modality.
Want et al. ~\cite{wang2016large} propose an online multitask learning and decision-making approach to coordinate machine actions for flexible manufacturing.

% Although MMMT learning has long been investigated in the joint area of computer vision and natural language processing, its applications in autonomous systems are still at early stage, especially with multimodal sensor fusion and multi-task followed by actuation and control.

\mysubsec{Neural architecture search (NAS).}
While NAS has been a great success in automatically searching for single machine learning models~\cite{elsken2019neural}, it is also being adopted in multi-modal or multi-task architecture search.
Yu et al.~\cite{Yu2020DeepMN} propose a deep multi-modal neural architecture search framework, namely MMNas, for various multi-modal learning tasks.
% first, an encoder-decoder based unified backbone is constructed from a pool of predefined operations; then, fixed task-specific heads are attached to tackle different multimodal learning tasks.
MFAS~\cite{PrezRa2019MFASMF} is another multi-modal fusion search approach by defining a superset of modern fusion architectures.
%and uses a sequential model-based optimization (SMBO) to search for a fusion architecture.
MTL-NAS~\cite{gao2020mtl} is a multi-task NAS work that tackles general-purpose multi-task learning
%which first builds multiple fixed single-task network branches, 
by searching for layer-wise feature sharing/fusing scheme represented by cross-branch edges.
Liang et al.~\cite{liang2018evolutionary} propose an evolutionary NAS for multi-task architecture search.
%by evolving hyperparameters, modules, and module routing topologies.
%starting from an initial soft ordering based architecture~\cite{meyerson2018beyond}.
However, NAS for MMMT architectures hasn't been studied, and no work has considered the hardware performance during the search.

\section{collaborative mmmt learning}
\label{sec:collab_mmmt_learning}

% \begin{figure}
%     \centering
%     \includegraphics[width=0.5\textwidth]{Figs/MMMT.pdf}
%     \caption{The difference between general MMMT learning and autonomous system specific MMMT learning, where the later requires historical system states as its additional modalities and priorities the control loss as its collaborative objective.}
%     \label{fig:MMMT}
% \end{figure}

\begin{figure}
    \centering
    \includegraphics[width=0.47\textwidth]{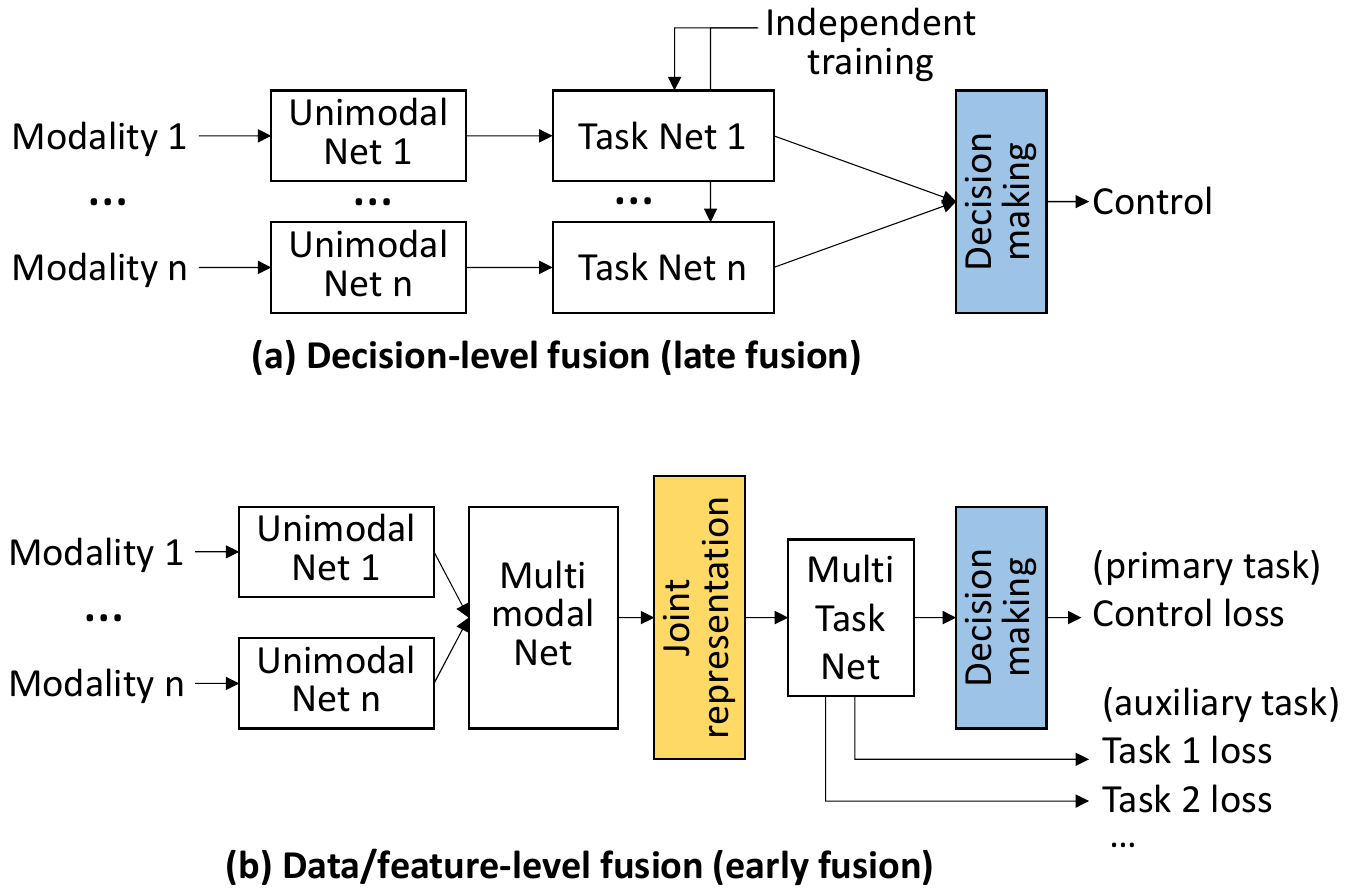}
    \caption{Decision-level fusion (late fusion) and data/feature level fusion (early fusion).}
    \label{fig:early_late_fusion}
    \vspace{-8pt}
\end{figure}

Although there are pioneer works that are exploring MMMT learning in autonomous systems, there are unique challenges that must be addressed.
Specifically, we discuss two salient differences of MMMT learning in autonomous systems: collaborative multimodal fusion, and collaborative hybrid multi-task learning.

\subsection{Collaborative Multimodal Fusion}

Collaborative learning is of great importance in autonomous systems, especially given distributed devices, multiple sensors, and multiple tasks that are potentially correlated to one another.
For example, swarm robotics need to interact with one another to collectively solve challenging problems such as pattern-formation and path-planning~\cite{cao201914}.
% Another example is the residential and commercial building management and control system, where the heating, ventilation, and cooling (HVAC) is adjusted to save energy without compromising the comfort of occupants, through incorporating an infrared (IR) camera with an optical (OP) camera to provide collaborative intelligence at low power and enhanced accuracy~\cite{cao2018smart}.
In autonomous factory, i.e., smart manufacturing, each camera or environment sensor captures only local information, and the global information can only be gained from a smart integration of all the sensors~\cite{Kusiak2018smart}.
In autonomous driving cars, sometimes large objects can only be detected and recognized from multiple cameras, whose input images are expected to be fused as early as possible.

In autonomous systems, there are still few works that have fully exploited the core technologies of multimodal learning, such as joint representation and modality translation~\cite{Baltrusaitis2019Multimodal}.
Instead, the majority of sensor fusion solutions in autonomous systems are still \textit{late fusion}~\cite{aranjuelo2018multimodal, huang2020multi, Vora2020PointPaintingSF, Yin2020FusionLaneMF, Nobis2019ADL, Nabati2020CenterFusionCR, brena2020choosing}, meaning that  after each of the modalities has made a decision, the system integrates the decisions using a fusion mechanism such as averaging, voting, AdaBoost, etc.
Although such mechanisms are simple and usually explainable, they may lose low-level interactions between the modalities.
Moreover, since they largely rely on empirical experiences and human expertise, 
%for large scale and complex autonomous systems with large numbers of multimodal sensors, where each sensor only captures a fraction of the overall environment, 
it is difficult for manually designed algorithms to make adequate decisions that need correct understanding of the overall environment.

Therefore, we propose to apply neural network based multimodal learning for autonomous system sensor fusion, in a collaborative way in an early stage. Fig.~\ref{fig:early_late_fusion} illustrates the difference between a decision-level fusion (late fusion) and data/feature-level fusion (early fusion). Different data modalities, i.e., the sensor inputs, are integrated together through shallow modality-specific networks, called modality nets; then, the features are concatenated and fed into a shared modality net to obtain a joint representation for feature extraction and downstream tasks.
More importantly, given the complexity of the collaborative modality fusion and modality net architecture, automatic search for multi-modal fusion and modality net design is necessary.

\subsection{Collaborative Hybrid Multi-task Learning}

While most autonomous systems contain multiple tasks, there is a fundamental difference between general MMMT learning and MMMT in autonomous systems:
the goal of general MMMT learning usually is to uniformly optimize each individual task, such as image segmentation and captioning, while exploring the mutual influences among tasks;
in autonomous systems, however, the ultimate goal is to optimize the final actuation/control, such as steering and motor values, while the intermediate tasks, such as perception and localization, collaboratively serve the final goal.

% With respect to system control, there are two large categories of methodology: the traditional model-based decision making (e.g., Markov decision process), and the ML-enabled end-to-end solution~\cite{schwarting2018planning}.
% As exampled in Fig~\ref{fig:early_late_fusion} (a), the traditional model-based decision making is usually applied after each task has reached a decision. The advantage is the smoothness and interpretability of the decisions, while the disadvantage is it easily converges to a local optimum solution.
% The end-to-end training incorporates partial tasks together and direct trains the ML-model directly using the final system outputs~\cite{tampuu2020survey, solomon2019hierarchical, merrill2019end}.
% The advantages are the capability of dealing with comprehensive and unseen situations, while the interpretability and lack of sufficient training data remain challenging.
% On the one hand, the unique requirement of autonomous systems that different tasks must collaboratively serve the final control favors end-to-end training; on the other hand, interpretability requirement favors model-based decision making.

On the one hand, general MMMT explores task relations and aims to better utilize the positive transfer while to avoid negative transfer, which requires independent training of each task~\cite{crawshaw2020multi}; on the other hand, end-to-end training in autonomous system focuses more on its final goal rather than intermediate tasks.
To leverage the advantages of both MMMT and end-to-end training, in this work, we propose collaborative hybrid multi-task learning for autonomous systems, as illustrated in Fig.~\ref{fig:early_late_fusion} (b): the primary task is the system actuation/control, while the auxiliary tasks are all intermediate tasks such as perception and localization.
The MMMT model will be trained using combined loss function, defined as follows, in an end-to-end manner:
\begin{equation}
    min: \sum_{c \in C} Control\_Loss_{c} + \lambda \sum_{t \in T} Task\_Loss_{t}
    \label{eq:hybrid_loss}
\end{equation}
where $C$ is the set of control values and $T$ is the set of all the tasks.

Similar to multimodal learning, the optimal MTL architecture should also be automatically searched via NAS technology.

\section{Software/Hardware Co-design}
\label{sec:sw-hw-codesign}

\begin{figure*}
     \centering
     \begin{subfigure}[b]{0.131\textwidth}
         \centering
         \includegraphics[width=\textwidth]{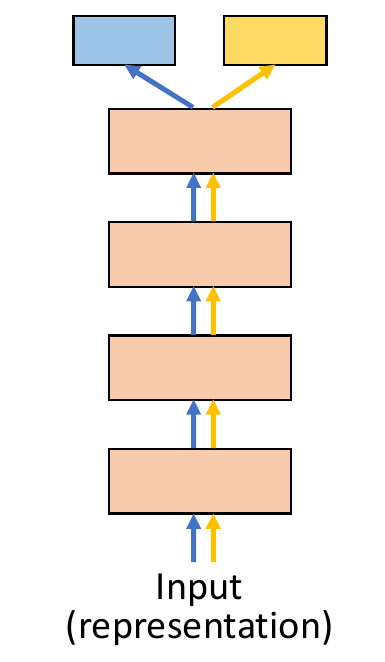}
         \caption{ \scriptsize Hard sharing~\cite{caruana1997multitask}}
         \label{fig:mmmt_1}
     \end{subfigure}
     \begin{subfigure}[b]{0.126\textwidth}
         \centering
         \includegraphics[width=\textwidth]{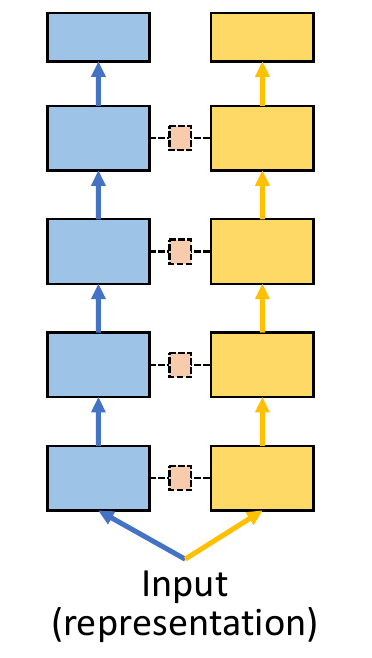}
         \caption{\scriptsize Soft sharing~\cite{duong2015low}}
         \label{fig:mmmt_2}
     \end{subfigure}
     \begin{subfigure}[b]{0.144\textwidth}
         \centering
         \includegraphics[width=\textwidth]{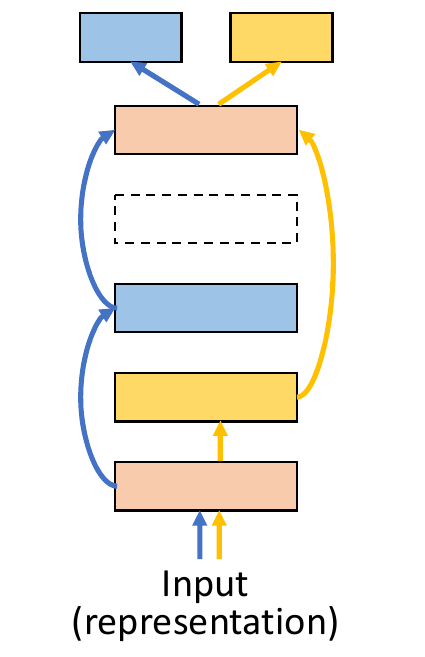}
         \caption{\scriptsize Adaptive sharing~\cite{sun2019adashare} }
         \label{fig:mmmt_3}
     \end{subfigure}
     \begin{subfigure}[b]{0.157\textwidth}
         \centering
         \includegraphics[width=\textwidth]{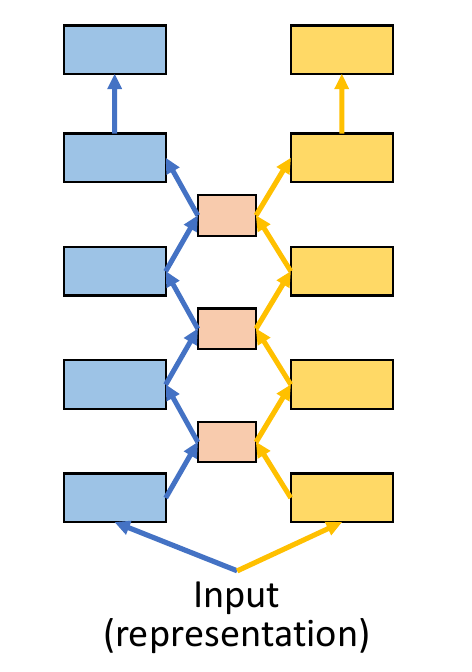}
         \caption{ \scriptsize Cross sharing~\cite{misra2016cross}}
         \label{fig:mmmt_4}
     \end{subfigure}
     \begin{subfigure}[b]{0.171\textwidth}
         \centering
         \includegraphics[width=\textwidth]{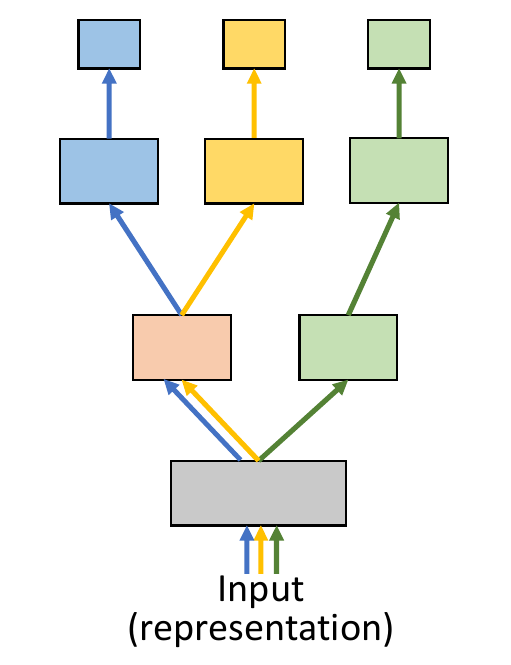}
         \caption{\scriptsize Branched sharing~\cite{lu2017fully}}
         \label{fig:mmmt_5}
     \end{subfigure}
     \begin{subfigure}[b]{0.21\textwidth}
         \centering
         \includegraphics[width=\textwidth]{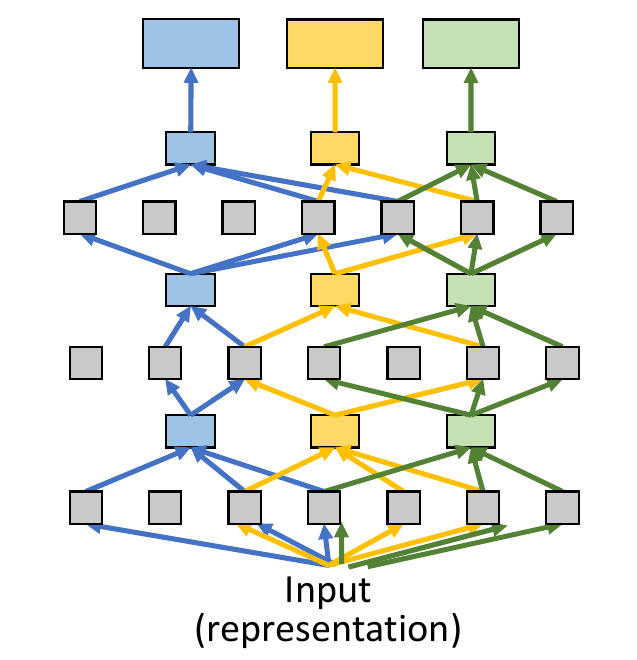}
         \caption{ \scriptsize Modular sharing~\cite{fernando2017pathnet}}
         \label{fig:mmmt_6}
     \end{subfigure}
        \caption{Different mechanisms of multi-task sharing, requiring different hardware mapping constraints and strategies. More details can be found in a multi-task learning survey~\cite{crawshaw2020multi}.}
        \label{fig:task_sharing}
\end{figure*}

\begin{figure*}
    \centering
    \includegraphics[width=0.9\textwidth]{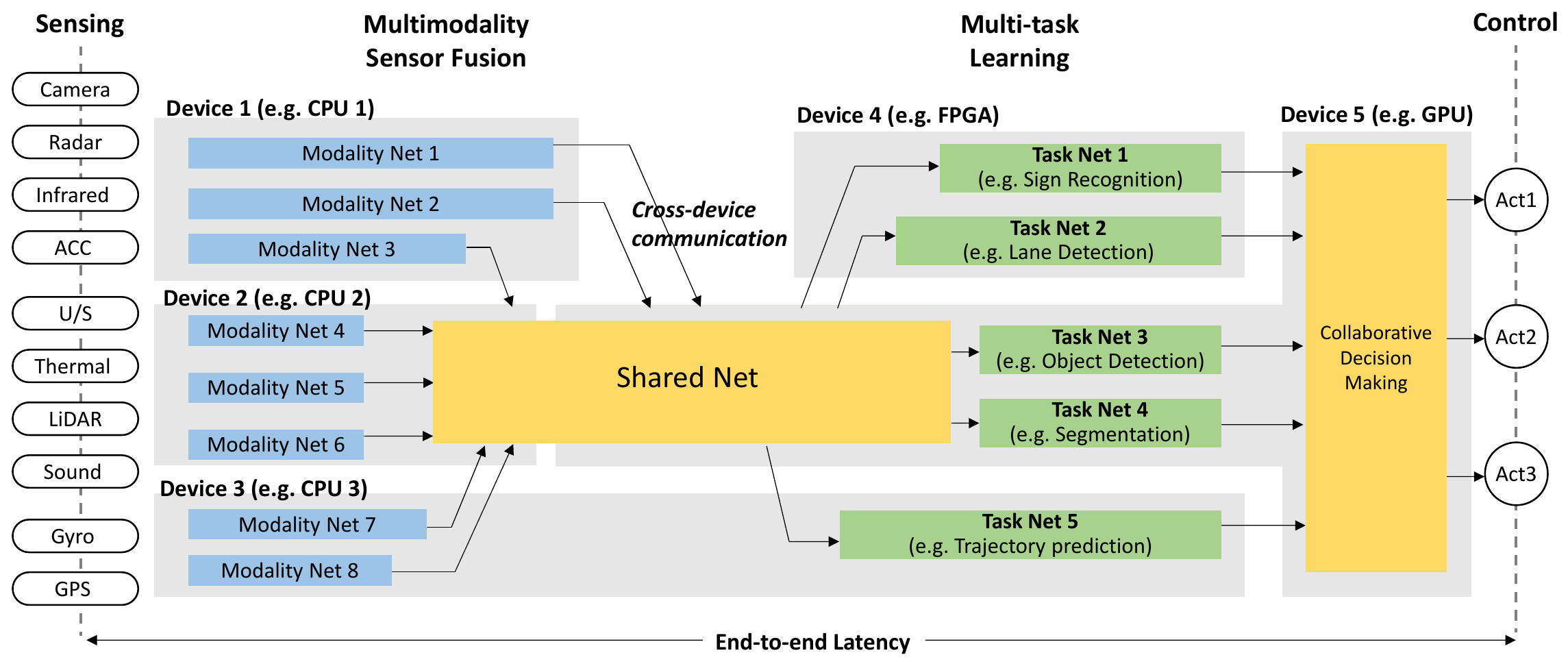}
    \caption{An illustration of an MMMT model and its mapping on a heterogeneous platform with multiple hardware devices. Ideally, minimizing the end-to-end latency requires similar completion time of all the tasks.}
    \label{fig:overall_mmmt}
\end{figure*}

While it is imperative to apply NAS technology for MMMT algorithm design, it is also indispensable to consider hardware deployment at the same time. 
% On the one hand, the complexity of a potential heterogeneous platform of autonomous systems significantly expends the hardware implementation solution space; on the other hand, the complexity of MMMT models further complicates the hardware implementation. 
Therefore, a software/hardware co-design approach for MMMT model and its hardware implementation design is of great importance.
We first discuss the challenges and opportunities of the co-design approach, and then formally formulate it as a multi-objective optimization problem, which can be solved by either continuous or discrete optimization techniques.

\subsection{Challenges and Opportunities}

Although there are many works exploring the architectural design for MMMT models, very few of them discuss the efficiency of hardware deployment, which is especially critical for autonomous systems. We discuss the following challenges and opportunities:
\begin{enumerate}
    \item {
The MMMT model structure greatly affects the overall computation complexity and the communication overhead. As illustrated in Fig.~\ref{fig:task_sharing}, various MMMT models have largely different hardware computation and communication complexity. For instance, Fig.~\ref{fig:task_sharing} (a) is the traditional hard sharing, where different tasks share the same set of network parameters and intermediate data, thus can significantly save the computation and communication across tasks. On the contrast, Fig.~\ref{fig:task_sharing} (b) to (e) illustrate other popular MMMT models, which introduce computation overhead (e.g., soft sharing and cross-talk sharing), communication overhead (e.g., cross-talk), or control overhead (e.g., adaptive sharing and modular sharing), and the hardware deployment efficiency may degrade significantly.
}
    \item {
Autonomous systems usually have heterogeneous platforms, that the MMMT model design is unaware of. For example, in an autonomous driving car, the computing platform may consist of GPUs, CPUs, FPGAs~\cite{hao2019hybrid}, deep learning accelerators, vision accelerators, image signal processors, and video encoders~\cite{nvidiaAGX}. Different mapping solutions from MMMT model to hardware components can result in largely different hardware performance. In addition to deep learning model execution, other overhead such as sensor data processing and cross-device communication is also critical to overall performance and thus cannot be overlooked.
Fig.~\ref{fig:overall_mmmt} shows an example of an overall software pipeline and its mapping on the hardware platform. Given the complexity of MMMT models and the heterogeneity of the hardware platform, the design space is extremely large. 
    }
    \item {
Another critical design metric for autonomous systems is the end-to-end latency, especially for life-critical decisions such as autonomous driving control.
As shown in Fig.~\ref{fig:overall_mmmt}, it is expected that the arrive time of the outputs of different tasks should be the same before the final decision making. Thus, the tasks that are not on the critical path can be allocated to less-powerful hardware components, while the tasks on the critical path must be allocated more carefully to reduce the end-to-end latency.
    }
\end{enumerate}

Therefore, for a fixed hardware platform, MMMT algorithms must be designed being aware of the hardware architecture and mapping strategy; for customizable hardware platforms, such as FPGA devices, both the MMMT model and the hardware implementation should be co-designed, and the system-level design solutions must be explored~\cite{yang2015system, chen2016platform}.
NAIS~\cite{hao2019nais} and EDD~\cite{li2020edd} are two existing automated neural architecture and implementation co-search frameworks. 
NAIS first highlights the necessity of a simultaneous DNN/implementation co-design methodology targeting both FPGAs and GPUs,
%, and takes the autonomous driving as an example use case;
while EDD is a fully simultaneous differentiable neural network architecture and implementation co-search framework.
%by fusing the DNN search variables and hardware implementation variables into one solution space.
While NAIS and EDD both demonstrate the success of software/hardware co-design, they are still in a single-model single-device fashion.
%the DNN is a single image classification task, while the target device is either a single CPU, GPU, or FPGA.

\subsection{MMMT model and Hardware Implementation Co-design}

Motivated by the necessity of software/hardware co-design, we propose automated MMMT model and heterogeneous hardware implementation co-search.
To enable efficient co-search, we must first 
formulate the problem in a way that:
1) the formulation shall involve both the algorithm design space and the hardware implementation design space;
2) different objective functions shall be included to provide QoR and QoS tradeoffs;
3) the formulation shall be solvable via optimization techniques such as discrete or continues optimization.

\mysubsec{Formulation.}
Inspired by NAIS and EDD, we formulate the software/hardware co-design of MMMT model and its hardware implementation as a joint design space including two sub-spaces:
the model architecture design space, denoted by $\mathcal{A}$ and parameterized by $\alpha$, and the hardware implementation design space on the heterogeneous platform, denoted by $\mathcal{I}$ and parameterized by $\omega$. 
%Typically, both $\alpha$ and $\omega$ are discrete values~\cite{elsken2019neural}.
The solution of $A_{\alpha}$ determines the QoR, while the solution of $I_{\omega}$ determines the QoS.
The formulation of $A_{\alpha}$ is similar to most existing NAS works such as DARTS~\cite{liu2018darts} and SNAS~\cite{xie2018snas}.
For the formation of $I_{\omega}$, we consider the following mapping on the heterogeneous platform that contains $D$ device components:
\begin{equation}
\begin{split}
& I_{w}=\{ m_{nn, d} \},~nn\in NN, d\in D \\
& s.t.: \forall nn\in NN, m_{nn, d} = \begin{cases}
1 & ~,\text{if}~nn~\text{runs on}~d \\  
0 & ~,\text{otherwise}
\end{cases} \\
\end{split}
\end{equation}
where $nn$ represents each MMMT model component, for example the modality net, the task-specific net, or the shared backbone net; $I_{\omega}$ is the collection of the mapping variables $m_{nn, d}$, either zero or one, depending on whether a component $nn$ runs on device $d$.

The goal of the co-design problem is twofold. First, it aims to search for an optimized MMMT model that can best leverage modality fusion and the positive transfer across tasks, i.e., one task can help improve the quality of another, to improve the software solution quality. Second, it searches for an optimized hardware mapping that each network component runs on the best hardware device with different computing and processing capabilities. Consequently, the overall heterogeneous autonomous system can not only achieve power reduction through backbone sharing but can also deliver high-quality AI models with well-balanced executions.
Therefore, we consider the following three objectives:
1) the software quality, measured as the hybrid loss function described in Eq.~\ref{eq:hybrid_loss};
2) the overall power consumption, measured as the power summation of all the active hardware components;
3) the end-to-end latency, measured as the longest path among the pairs of input modalities and output tasks.

Formally, we formulate the MMMT model and hardware implementation co-search problem as follows:
\begin{equation}
\begin{split}
    min:~ & Loss_{sw}(A_{\alpha}) + \gamma_1 \cdot Loss_{hw}(I_{\omega}) \\
    s.t.~ & Loss_{hw} = max\{ Lat_{xp(m, t)} | \forall m\in M, t\in T\} \\
    & ~~~~~~~~~~ +  \gamma_2 \cdot \sum_{d\in D }Power(d)
\end{split}
\label{eq:formulation}
\end{equation}
where $m$ is the input modality in the modality set $M$, $t$ is the specific task output in the task set $T$;
$d$ is the hardware device component that is active in the final deployment; $Power(d)$ is its power consumption;
$xp(m, t)$ is the path from modality $m$ to task $t$;
$Lat_{xp(m, t)}$ is the path latency from input modality $m$ to task $t$ on the hardware platform.

\mysubsec{Solutions.}
The formulation defined in Eq.~\ref{eq:formulation} can be solved via either discrete or continuous optimization~\cite{parker2014discrete, kochenderfer2019algorithms}.
For instance, evolutionary algorithms are commonly used in NAS, i.e., solving $A_{\alpha}$, through discrete optimization~\cite{liu2020survey}, while RNN and reinforcement learning~\cite{baker2016designing, zoph2016neural} are also widely adopted.

Another category of prevailing approach is to relax the discrete distribution to make it differentiable and then adopt continuous optimization techniques such as gradient descent.
Some existing NAS works have proposed such continuous relaxation to solve $A_{\alpha}$ through weighted sum~\cite{liu2018darts} or reparametrization and statistic sampling~\cite{kingma2014auto, maddison2017concrete, wu2019fbnet}.
Similarly, to solve $Loss_{hw}(I_{\omega})$ by continuous relaxation, first, the non-differentiable maximum operation among all paths can be relaxed using softmax.
Second, the discrete optimization of $I_{\omega}$ can use similar reparametrization and statistic sampling technology as follows:
\begin{equation}
\begin{split}
    Lat (\omega, \phi) & = \mathbb{E}_{XP \sim P_{\phi}(xp)} [ f_{\omega}(XP) ]  \\ 
    & = \mathbb{E}_{Z \sim q(z)} [ f_{\omega}( g_{\phi}(Z) ) ]
\end{split}
\label{eq:reparam}
\end{equation}
where $\phi$ is the original distribution parameter.
$q(z)$ is a fixed distribution (e.g., Gaussian, Gumbel~\cite{maddison2014sampling}).
$g_{\phi}(z)$ is the transformation function, where $z$ is sampled from distribution $q(z)$, and $x$ is computed from $z$ through the deterministic transformation $x= g_{\phi}(z)$; this is the reparametrization technique.
$f_{\omega}(\cdot)$ is the latency function under mapping parameter $\omega$. 
In this way, by doing continuous relaxation, the objective function is differentiable with respect to the sampling parameters, which thus can be optimized through continuous optimization technologies such as gradient descent.

By formulating the MMMT model and implementation co-design as a multi-objective optimization problem (Eq.~\ref{eq:formulation}), which can be solved either by discrete optimization or continuous optimization via sampling and reparametrization (Eq.~\ref{eq:reparam}), the co-design problem can be systematically studied and solved.

\section{conclusions and future directions}
\label{sec:conclusion}

In this paper we discussed the opportunities and challenges of exploiting multi-modal multi-task learning techniques in autonomous systems. 
The unique challenges that must be solved are collaborative multi-modal learning and collaborative hybrid multi-task learning.
We also discussed the software/hardware co-design opportunities and the necessity for MMMT in autonomous systems, especially targeting heterogeneous hardware platforms with a complex MMMT model. We formally formulated the MMMT and hardware implementation co-design as an optimization problem, and discussed the potential solutions through statistic sampling and reparametrization techniques. We advocate for further explorations of MMMT in autonomous systems and software/hardware co-design solutions.

\bibliographystyle{IEEEtran}
\footnotesize
\bibliography{ref}

\end{document}